\documentclass[conference]{IEEEtran}
\IEEEoverridecommandlockouts
\usepackage{cite}
\usepackage{amsmath,amssymb,amsfonts}
\usepackage{algorithmic}
\usepackage{graphicx}
\usepackage{textcomp}
\usepackage{multirow}
\usepackage{slashbox}
\usepackage{booktabs}
\usepackage{wrapfig}
\usepackage{titlesec}
\usepackage{subcaption}
\def\BibTeX{{\rm B\kern-.05em{\sc i\kern-.025em b}\kern-.08em
    T\kern-.1667em\lower.7ex\hbox{E}\kern-.125emX}}
\IEEEoverridecommandlockouts

\IEEEpubid{\makebox[\columnwidth]{978-1-6654-3902-2/21/\$31.00~  
\copyright~2021 IEEE \hfill} \hspace{\columnsep}\makebox[\columnwidth]{}}

\usepackage{soul}

\usepackage{xcolor}
\def\BibTeX{{\rm B\kern-.05em{\sc i\kern-.025em b}\kern-.08em
    T\kern-.1667em\lower.7ex\hbox{E}\kern-.125emX}}
\begin{document}

\newcommand{\mt}[1]{\textcolor{red}{\# #1 (MT) \# }} 
\newcommand{\ra}[1]{\textcolor{blue}{\# #1 (RA) \# }}
\newcommand{\minh}[1]{\textcolor{brown}{#1}} 
\newcommand{\minhnote}[1]{\textcolor{magenta}{\# #1 (MV) \# }} 

\title{Learning Interpretation with Explainable Knowledge Distillation\\

}

\author{\IEEEauthorblockN{Raed Alharbi \hspace{12pt} Minh N. Vu \hspace{12pt} My T. Thai}
\IEEEauthorblockA{\textit{Computer and Information Science and Engineering Department} \\
\textit{University of Florida, Gainesville, FL, USA }\\
\{r.alharbi, minhvu, mythai\}@ufl.edu}
}

\maketitle

\begin{abstract}

Knowledge Distillation (KD) has been considered as a key solution in model compression and acceleration in recent years. In KD, a small student model is generally trained from a large teacher model by minimizing the divergence between the probabilistic outputs of the two. However, as demonstrated in our experiments, existing KD methods might not transfer critical explainable knowledge of the teacher to the student, i.e. the explanations of predictions made by the two models are not consistent. In this paper, we propose a novel explainable knowledge distillation model, called XDistillation, through which both the performance the explanations' information are transferred from the teacher model to the student model. The XDistillation model leverages the idea of convolutional autoencoders to approximate the teacher explanations.   Our experiments shows that models trained by XDistillation outperform those trained by conventional KD methods not only in term of predictive accuracy but also faithfulness to the teacher models.

\end{abstract}


\section{Introduction}

With the extensive deployment of deep neural networks models (DNNs) on lightweight and low computing resources, such as mobile devices, or Internet-of-Thing (IoT) devices, Knowledge Distillation (KD) has been shown as one of the most promising approaches to transfer knowledge from a large model, called teacher, to a smaller one, called student, without loss of predictive power and validity \cite{44873, han2015learning}. The core concept is that the teacher model is utilized during a KD process to guide the student model by passing on substantial information. With similar performance, the student models have much less parameters and can be deployed on less powerful hardware. KD has been successfully used in several application domains, such as computer vision and natural language processing \cite{cho2019efficacy}.


With an increasing alarm by the public and researchers on the lack of interpretablity of current DNNs, that is, they have been used as black-boxes with a little explanation, transferring the explainability is as important as preserving the predictive accuracy in knowledge distillation. Being able to adopt the teacher explanations is significantly desirable because: 1) Explanations provide transparency to the models' prediction, thereby increasing trustworthiness in using models \cite{alharbi2021evaluating}. 2) Trustworthiness and faithful explanations can identify models' failures and bias when not all possible scenarios are testable~\cite{arrieta2020explainable}, thereby avoiding several shortcuts learning that existing DNN models have been exhibiting \cite{Geirhos2020}.

\begin{figure}[h]
\centering
    \includegraphics[width=1\columnwidth]{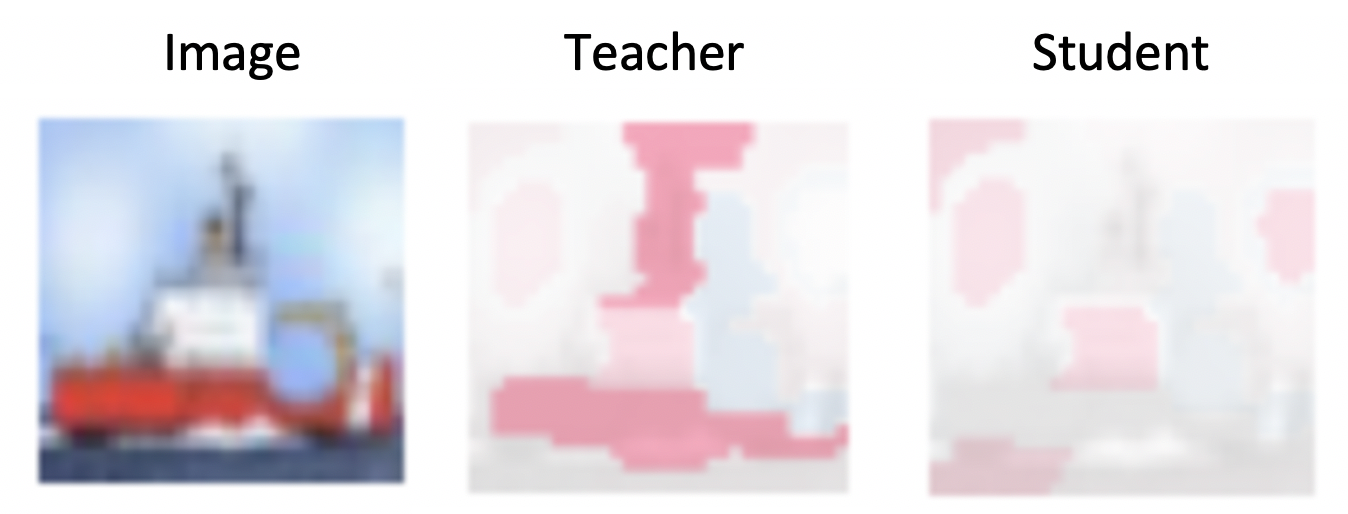}   
    \caption{The inconsistency between the explanations of the predictions of the two models.}
    \label{non_cons}
\end{figure}

Unfortunately, while existing KD approaches have been focused on preserving the performance accuracy, the explanation on why the model makes its prediction is not transferred from the teacher to the student model. As most of the existing teacher models are not interpretable by themselves, we use a post-hoc explainer SHAP \cite{lundberg2017unified} to explain the important features to the model's prediction. Figure \ref{non_cons} illustrates an inability of transferring the explanation knowledge of existing KD models. As can be seen in Figure \ref{non_cons}, the explanations of the teacher and student models of the same image are inconsistent, using SHAP.





Along this direction, this paper proposes an effective KD model, called XDistillation, which not only maintains the teacher's performance but also approximates the teacher's explanation. Sitting in the heart of XDistillation is a novel explainable features fusion technique that significantly reduces the total number of student parameters while ensuring consistency between teachers' and students' explanations. Extensive experiments demonstrate that XDistillation provides consistent explanations between the teacher and student models while being on par with existing KD techniques in terms of performance accuracy. 


The  remainder  of  the  paper  is  structured as  follows.  Section  \ref{related} presents  the  related works. Our proposed XDistillation model is introduced in  Section \ref{model_details} while the experimental analysis are discussed in Section \ref{exp_details}. Finally, Section \ref{sum_up} concludes our paper.


\begin{figure*}[h]
\centering
    \includegraphics[width=2.1\columnwidth]{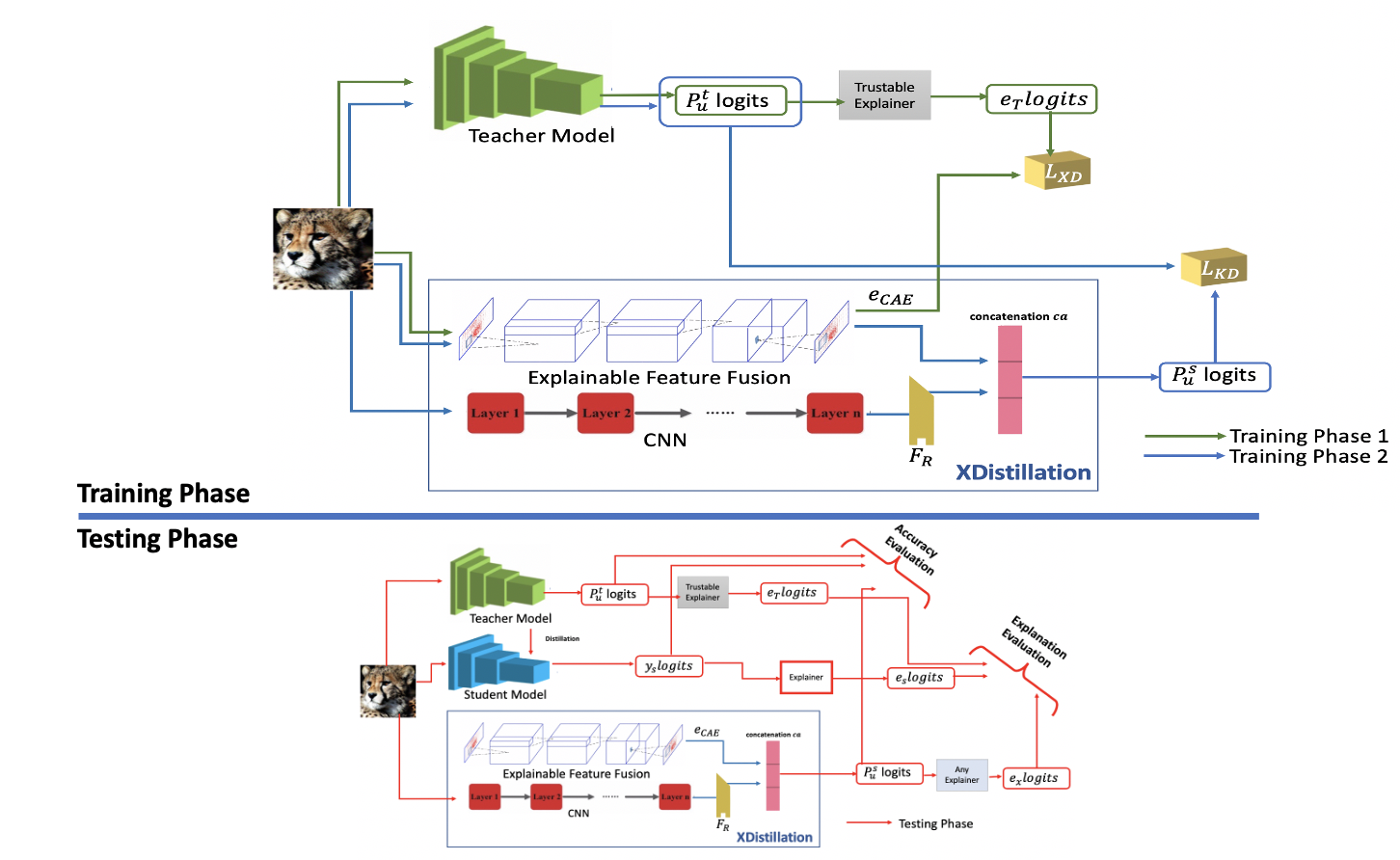}   
    \caption{The overall architecture of XDistillation. The top network illustrates the training phases using two losses functions as described with more details in section \ref{model_details}. The first loss minimizes the inconsistency between the explanations of a teacher model and our approximated explanations by a novel explainable feature fusion technique. The second loss minimizes the predictions among the two models. The bottom network depicts the evaluation measures in term of the accuracy performance and explanations.} 
    \label{i_model}
\end{figure*}

\section{Related Work}
\label{related}

\textbf{Knowledge Distillation.}
The concept of knowledge distilling relates to the idea of model compression, where a teacher model guides the student model while retaining a high degree of generality \cite{44873}. Since then, there have been studies on various methods of knowledge distillation to improve the student model. Zagoruyko \emph{et al.} in \cite{zagoruyko2016paying} proposed the use of attention maps where 
Heo \emph{et al.} \cite{heo2019knowledge} utilize the activated neurons to distill only the valuable information to a student model, and Tung \emph{et al.} \cite{tung2019similarity} follow activation patterns of neurons between a teacher and student model. 
Yet, no prior attempts have been made to distill the explanations from a teacher a student to improve both a student's performance and explanations.


\textbf{Explainable Artificial Intelligence (XAI).}
\label{ex_AI}
One of the major deficiencies to the deployment of DNN models is the lack of transparency \cite{vu2020pgm}. The black-box structure of these models permits strong predictions. However, they cannot be explained explicitly. This problem has initiated a new discussion about explainable AI \cite{lipton2018mythos}. The purpose of employing interpretability in AI models is to verify  the model predictions and the important features that contribute to these predictions can be interpretable to end-users \cite{lipton2018mythos, vu2020pgm} whereas preserving a high degree of accuracy performance in the same time.

Two popular types of methods have been presented in the literature to tackle the interpretability needs: (1) backpropagation-based such as CAM \cite{zhou2016learning} and GradCAM \cite{Selvaraju_2017_ICCV}. (2) perturbation-based such as occlusion analysis \cite{fong2019understanding}, LIME \cite{ribeiro2016should}, and SHAP \cite{lundberg2017unified}. On the one hand, predictions and explanations in backpropagation-based methods are both generated by the same fundamental technique. Those methods use the properties of Convolutional Neural Networks (CNNs) by utilizing the activations of the models convolution layers to describe the relation of the input to the output \cite{Selvaraju_2017_ICCV}. On the other hand, the perturbation-based methods address the impact of the input perturbations on the output to identify important features for explanations.

\textbf{Autoencoders (AEs)} Using an autoencoder is a common technique for ensuring that the input and output are as comparable as possible \cite{wang2016auto}. The autoencoder is a form of neural network that is capable of learning a compressed version of the input data in unsupervised manner. Autoencoders have been made significant contributions to the field's application and research including dimensionality reduction \cite{wang2016auto}, image improvement \cite{lore2017llnet} and detection of outliers \cite{chen2017outlier}. Recently, different variant of AEs has been proposed such as convolutional autoencoders (CAE) \cite{masci2011stacked} and denoising autoencoders (DAE) \cite{ashfahani2020devdan}. One of the primary benefits of AEs is their ability to approximate any function. However, no attempt has been made previously to approximate the explanations of model predictions

In this paper, we select well-known explainers from backpropagation-based and perturbation-based categories to use them during the training and testing phases, as it describes later in Sections \ref{model_details} and \ref{exp_details}. Particularly, for the first category, GradCAM is chosen as it is computationally efficient (requiring only a single forward and backward pass via a network). For a particular decision, GradCAM  gives a value to each neuron utilizing gradient information that flows into the CNN’s last convolutional layer \cite{Selvaraju_2017_ICCV}. LIME and SHAP are selected from the second category as providing users with explanations in the form of feature importance score has been seen as a successful technique \cite{ribeiro2016should,lundberg2017unified}.


\section{The Proposed Model}
\label{model_details}

We design a simple but effective model, called XDistillation, which provides explanation-distillation from the teacher to increase the performance and enhance the explainability of the student. Figure \ref{i_model} shows an overview of XDistillation where the key functionalities are explainable features fusion component, how the interpretable features output of this fusion component is fed into Convolutional neural networks (CNNs) via concatenation layers, and how to constrain the signal outputs from CNN (feature reduction-$F_{R}$ in Figure \ref{i_model} ) to allow more explainability freedom from the explainable fusion part.

Due to an importance of the explainable features fusion component, we first describe it in section \ref{Ex_F_F}. 
Section \ref{ex_model_details} presents the overall picture of the distillation technique, including the feature reduction in a CNN structure, concatenation layers, and the details of the training phases. 

\begin{figure}[h]
\centering
    \includegraphics[width=0.65\columnwidth]{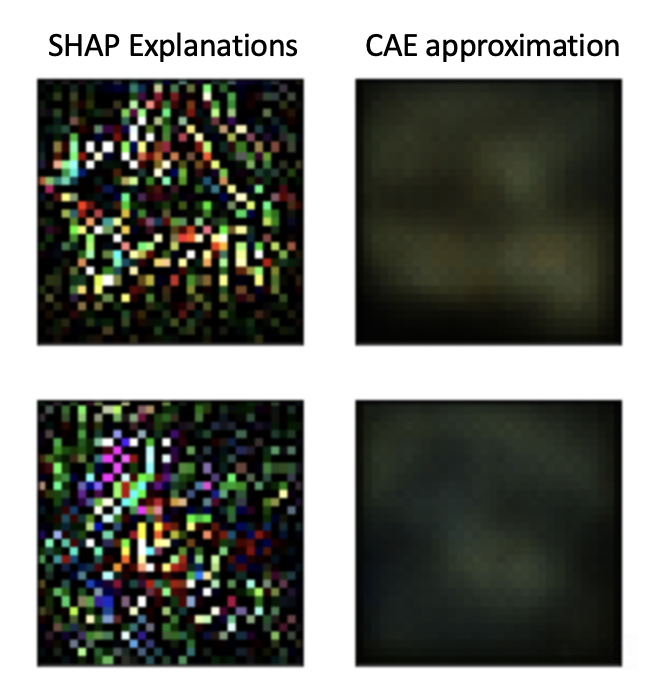}   
    \caption{ The output of convolutional autoencoder  after feeding the SHAP explanation. The CAE trained on the original SHAP representation. We can observe the sparsity issue on the left images. Hence, the CAE fails to mimic the desired explanations.
   }
    \label{cae_fail}
\end{figure}

\begin{figure*}[h]
\centering
    \includegraphics[width=1.7\columnwidth]{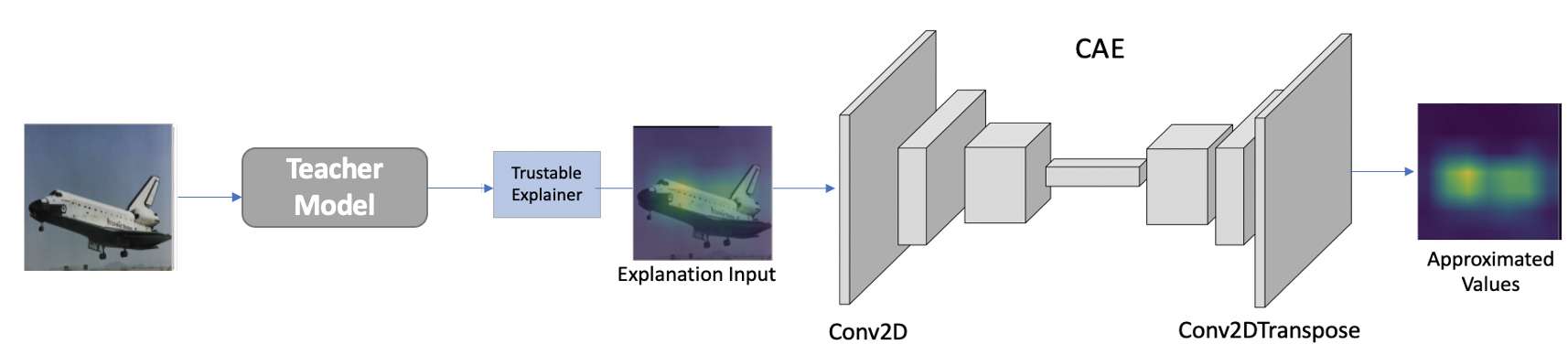}   
    \caption{The overall architecture of the explainable features fusion component. Given an explanation image as input, the fusion component generates the approximated explanation, consisting two main functionalities: (1) a new representation for explanations and (2) an approximation of an image explanations using convolutional autoencoder.}
    \label{Fea_Fusion}
\end{figure*}

\subsection{Explainable Features Fusion} 
\label{Ex_F_F}

The goal of the explainable features fusion component is to first extract the explanation features from a teacher model and then approximate those valuable features. The explainable fusion component includes a reliable explainer and CAE as illustrated in Figure \ref{Fea_Fusion}.

Since existing teacher models work in a black-box manner without explicitly explaining why they make their decisions, we will use post-hoc explainers, such as SHAP, LIME, GradCAM, to extract important features that contribute to the model's output prediction, called explanations. Next,  we leverage the idea of convolutional autoencoder (CAE) \cite{masci2011stacked} to learn and transfer the explanations of the teacher to the student. 

Unfortunately, the current representation of explanations from existing explainers such as SHAP, LIME, GradCAM is not ready for us to feed into CAEs. Furthermore, CAEs were designed to imitate its images input as nearly as possible to its output, not to approximate and extract important features to predictions with the lowest possible number of parameters. We will address these two challenges in sections \ref{shap_Rep} and \ref{cee_section} accordingly.


\begin{figure}[h]
\centering 
    \includegraphics[width=0.8\columnwidth]{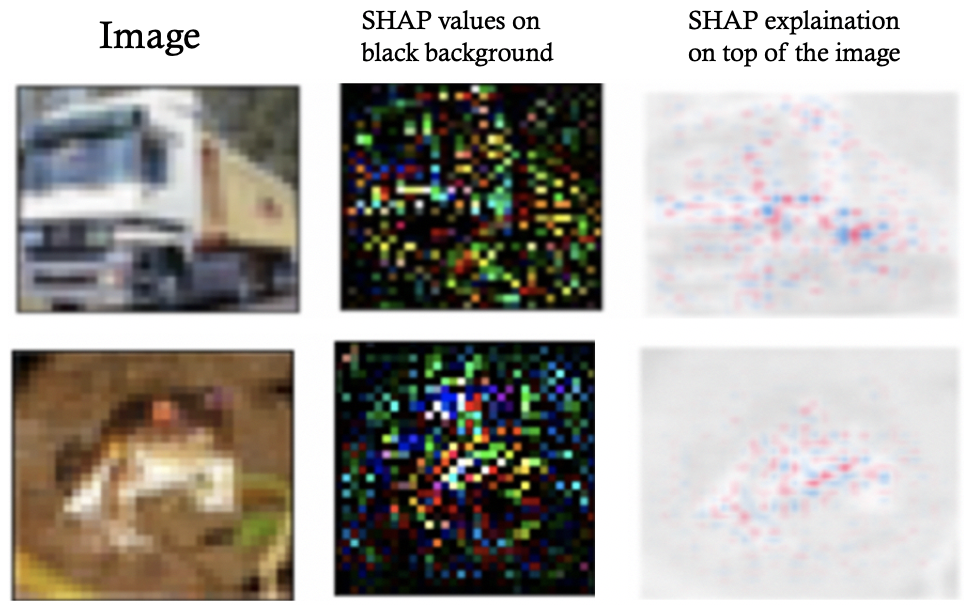}   
    \caption{The sparsity issue of SHAP values. We can observe the sparsity issue between RGB channels for SHAP explanations of the two images. Also, the SHAP explanation of the truck image on the third column does not reveal clear information whether the used model classified the image correctly since the positive and negative values are mixed.} 
    \label{shap_spa}
\end{figure}

\begin{figure}[h]
\centering
    \includegraphics[width=0.8\columnwidth]{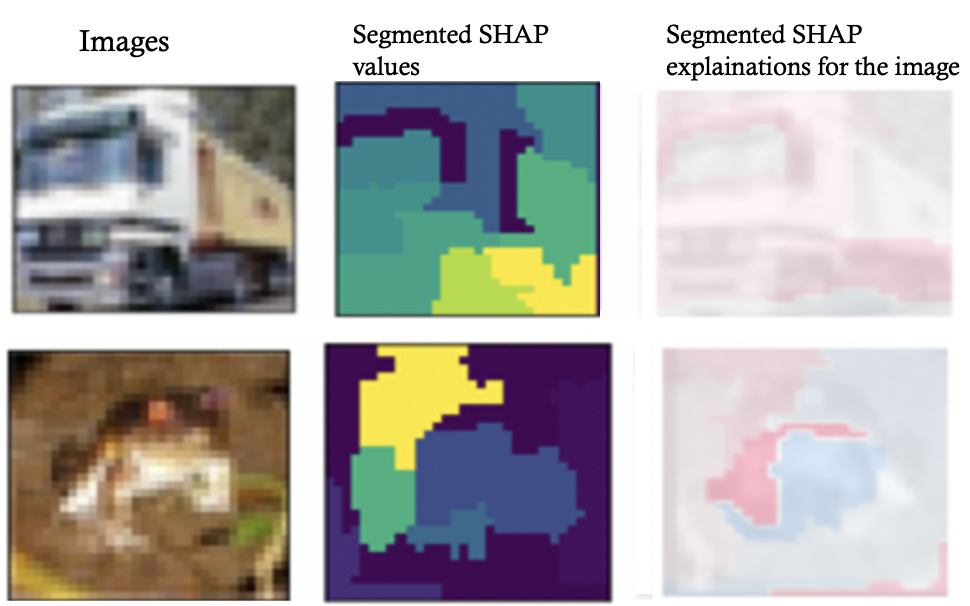}   
    \caption{The new representation of SHAP values. We can observe the improvement of the SHAP explanation of the two images using the new representation. Moreover, we can visually say that the truck image on the third column is classified correctly as the SHAP values react positively with the truck in the image.} 
    \label{shap_seg_}
\end{figure}

\vspace{8pt}
\subsubsection{New Representation for Explanations} 
\label{shap_Rep}
\vspace{8pt}

For the sake of simple demonstration, we will use SHAP as an explainer for our model as SHAP is one of the state-of-the-arts among existing explainers \cite{lundberg2017unified}. In our experiments, we will demonstrate the use of our new explaining representation for other explainers.

The current SHAP representation cannot be approximated by CAEs due to a sparsity issue, as shown in Figure \ref{cae_fail}. The sparsity issue occurs when the available explanations among the channels are not enough for the CAE to mimic the explanations. Therefore, the CAE fails because decisions of filling those empty areas with SHAP explanations (positively or negatively) cannot be taken randomly. In particular, SHAP explainer does not work smoothly with RGB (Red, Green, and Blue) images. The issue is far different for RGB images in comparison to grayscale images. Figure \ref{shap_spa} shows SHAP explanation on top of trained VGG16 model \cite{SimonyanZ14a} for a random image from CIFAR-10 dataset \cite{krizhevsky2009learning}. As can be seen, the SHAP values are quite sparse and spread between the channels. For each channel, most values are zeros, making the issue like a high-dimensional classification problem. Additionally, the non-zero values are quite high (e.g., $-5$, $+3$, etc). In other words, a sparse representation of data is a representation in which few parameters or coefficients are not zero, and many are (strictly) zero. 


To overcome the sparsity issue while preserving this XAI goal, we divide an image into smaller superpixels, given the SHAP values. For each superpixel patch, we sum all SHAP values within that patch. The superpixel in this way will describe an actual effect of SHAP values that we will have on the model's output (either negative or positive). Figure \ref{shap_seg_} shows the new representation which solves the sparsity issue. The two images in Figure \ref{shap_seg_} were classified correctly by training the VGG16 model. Then, the SHAP explanations of the two images segmented with different patches (images in the second column). The new representation of SHAP in the third column clearly demonstrates different dominant areas in the image either positively or negatively, reflecting the model's confidence in classifying those images. For instance, the model was certain that the object in the first image is a truck, as we can see from the positive SHAP values for most patches. However, the model doubts the object in the second image to be a fork. 

\begin{figure}[h]
\centering
    \includegraphics[width=0.6\columnwidth]{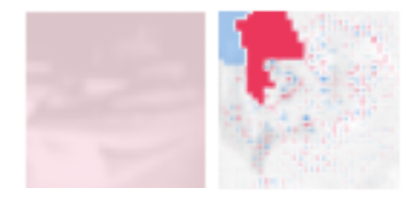}   
    \caption{ Non adequate SHAP segmentation.}
    \label{non_ad}
\end{figure}

In order to do a proper superpixel segmentation to handle the sparsity issue, we adopt a simple linear iterative clustering (SLIC) algorithm \cite{noh1976slic}, which is a local $k$-mean cluster of 5D pixels. SLIC takes the desired number of approximately equally-sized superpixels $k$ as input and clusters them using a new distance measure that takes superpixel size into account. 

Selecting a value $k$ is important. By experiment, $k$ should be in a range of $[3, 20]$. If $k < 3$, we will have large patches for an image as it is shown in the left image in Fig. \ref{non_ad} which merge the SHAP positive and negative values, and therefore we cannot know what parts contribute to which areas. On the other hand, if $k > 20$, we will have the same sparsity issue for some parts of the image as shown in the right image of Fig. \ref{non_ad}. In our experiments shown in Section \ref{exp_details}, we set $k = 19$.

\vspace{8pt}
\subsubsection{Convolutional Autoencoder (CAE)}
\label{cee_section}
\vspace{8pt}

{\bf Preliminaries and Notations.} Generally, convolutional autoencoders  (CAEs) are a form of neural networks that consist of two parts: (1) encoder, which learns to encode the important information of an input into compressed latent representation. (2) decoder, which reconstructs the input again from the latent representation \cite{masci2011stacked}. Mathematically, CAE maps an input $x \in R^d$ to a latent representation $h \in R^l$, where usually $d > l$. The latent representation of the $i$-th feature map for a single-channel input $x$ is given by

\begin{equation}
\label{latent}
h^{i}=\sigma\left(\mathrm{x} * \mathrm{~W}^{\mathrm{i}}+b^{i}\right)
\end{equation}
where the bias $b$ is applied to the entire map, $\sigma$ is an activation function, $W$ is the trainable weights and $\ast$ denotes 2D convolution layer. 




{\bf Current Usages and Limitation.} The most popular use of CAE is feature extractor for classification \cite{masci2011stacked}. This can be done in two ways: (1) Eq. (\ref{latent}) is used to preserve the most important information from an input via a convolution layer. Then, the latent representation is used with different CNNs for a classification task. (2) the encoder (Eq. (\ref{latent})) and decoder (reconstruction in CAEs \cite{masci2011stacked}) parts are used, but with replacing some of the convolution layers in the decoder part with fully connected layers to do the classification task.

Since the primary goal of using CAEs is not to do the classification task, most of current CAEs use the fully connected layers as part of their frameworks which increase the total number of parameters. If naively applying this approach, the number of parameters in a student model significantly exceeds that of a teacher model. Also, using existing modified CAEs to do the classification task requires us to use Softmax on the last layer, which makes us lose critical information of explanations.

{\bf Our Proposed CAE.} To overcome the limitations, we first adopt the convolutional autoencoder as an image transformation function. The input of the auto-encoder $x$ will be an image, and instead of a reverse mapping function, its purpose is to approximate the SHAP explanations of the teacher model $t(x)$, which will be a convolutional neural network. Our objective here is to reconstruct (approximate) SHAP explanation of the teacher model as it is shown in Figure \ref{Fea_Fusion}.

We use Eq. (\ref{latent}) to calculate the latent representation of the $i$-th feature map for a single-channel input $x$.
The estimation of the SHAP values $e_{T}$ of $t(x)$  (the first part of the training phase in Figure \ref{i_model}) is approximated as follows: 
\begin{equation}
\label{CAE}
e_{CAE}=\sigma\left(\sum_{i \in H} \mathrm{~h}^{\mathrm{i}} * \tilde{\mathrm{W}}^{\mathrm{i}}+b\right)\end{equation}

where $\tilde{\mathrm{W}}$ denotes the flip operations over the dimension weights and $H$ is a set of latent feature map.




We use different rule of thumbs to preserve the parameters constraint for teacher-student model while allowing CAE to approximate SHAP explanations successfully at the same time. First, the non-linearity data processing is the primary goal of an activation function after each layer \cite{ranzato2007unsupervised}. Because convolution/deconvolution operations are multiplications in nature, our visualization revealed that the result of the convolution/deconvolution operations (the values of feature maps) increased sharply from layer to layer, preventing the CAE model from converging during training. Thus, the use of hyperbolic tangent activation functions, which constrain the resulting values of feature maps to the interval [-1,1], sets appropriate limits on the values of feature maps at the end of the decoder part as well as maintaining the SHAP valuable information since SHAP values will be within the range for a model that outputs a probability \cite{NIPS2017_7062}, and provides good convergence of the whole model. 

Second, the basic structure of the autoencoder is extended by changing the completely connected layers to convolution layers in the convolutional autoencoder because (1) leveraging convolutional operations not only for slower training time and chances of reducing overfitting but also maintaining a reasonable total number of parameters. For instance, the total number of parameters for the small CAE version in Table \ref{cae_hp} (Section \ref{exp_details}) are reduced from $279,872$ to only $1077$. (2) The 2D image structure is ignored by fully connected AEs or partially by fully connected layers in CAEs. This not only creates redundancy in the parameters while handling inputs of exact magnitude.

A sample of successful SHAP approximation by the fusion component is shown in Figure \ref{Fea_Fusion} where SHAP explainer is first employed on top of a correct prediction of the teacher model for a plane image. The new representation of SHAP values is then prepared as described in Section \ref{shap_Rep} and passed to our new CAE, which produces the explainable features that describe the plane visually, as it can be seen in Figure \ref{Fea_Fusion}.

\subsection{XDistillation Model }
\label{ex_model_details}

We are now ready to describe the rest of our proposed XDistillation model. Figure \ref{i_model} demonstrates the overall architecture of XDistillation model, which consists of (1) explainable feature fusion component, and (2) CNN model. We use the first component, our CAE model, to transfer the explanations knowledge from a teacher model to the CNN model. Additionally, we use the feature reduction $F_R$ to allow more explainabiilty freedom associated with the first component.

For a given image $x$, a teacher model generates a vector of prediction scores

$$\mathbf{s}^{t}(x)=\left[s_{1}^{t}(x), s_{2}^{t}(x), \ldots, s_{u}^{t}(x)\right],$$
where $t$ refers to the teacher model, $x$ is the input and $u$ is the value of the scores which are then transformed to probabilities:

$$p_{u}^{t}(x)=\frac{\exp^{s^{t}(x)}}{\sum_{j} \exp^{s_{j}^{t}(x)}}.$$ 

Since the trained neural networks generate probability distributions with peaks that might be less instructive, we utilize the method in  \cite{44873} for temperature scaling to "Smooth" those probabilities:

\begin{equation}
\tilde{p}_{u}^{t}(x)=\frac{\exp^{s_{u}^{t}(x) / \tau}}{\sum_{j} \exp^{s_{j}^{t}(x) / \tau}},
\end{equation}
where $\tau > 1$ is a temperature hyperparameter. 

Similarly, the CNN model inside the XDistillation model in Figure \ref{i_model} also returns a smooth output denoted by $\tilde{\mathbf{p}}^{s}(x)$ where $s$ refer to the CNN model.


Figure \ref{i_model} illustrates $2$ training phases of XDistillation. In the first training phase, SHAP explainer is used to obtain ground truth explanations of $t(x)$. Following that, We train CAE using the Mean Absolute Error (MAE) as follows ($L_{XD}$ loss in Figure \ref{i_model} ): 

\begin{equation}L_{XD}=\frac{1}{2 n} \sum_{i=1}^{n}\left(e_{CAE_i}-e_{T_i}\right),
\end{equation}
where $e_{CAE_i}$ is described in Eq. (\ref{CAE}) and $e_{T}$ is the ground truth explanations of $t(x)$. The weights are then updated with classic backpropagation with stochastic gradient descent.

With the trained CAE model, we feed its output to a CNN; the resulting representations are flattened and concatenated before they are used as an input of the classifier part of the model. An illustration of our architecture can be found in Figure \ref{i_model}. 

In the second training phase, we use the knowledge distillation loss ($L_{KD}$ loss in Figure \ref{i_model}) which can be defined as follows:

\begin{equation}
\mathcal{L}_{K D} =-\tau^{2} \sum_{b} {p}_{u}^{t}(x) \log {p}_{u}^{s}(x)
\label{eq:kd}
\end{equation}

We extended the objective function with feature reduction $F_R$ for the last layer after the concatenation $ca$. In combination with the knowledge distillation loss defined in Equation \ref{eq:kd}, this yields the total loss function:

\begin{equation}
\label{eq:total_loss}
\mathcal{L} = \mathcal{L}_{c l s}+(1-\alpha) \mathcal{L}_{K D} +  F_{R}( \lambda ||c||^2_2 )
\end{equation}

where $\lambda$ is signal controller hyperparameter, and $c$ is the weight of last layer of the concatenation part. 
The feature reduction $F_{R}$ adds coefficient squared magnitude to the loss function as penalty term in last layer (concatenation layer as it is shown in Figure \ref{i_model}). The cross entropy function is defined as: 

$${L}_{cls} = -\sum_{a=1}^{M} y \log(p(y_a))$$

where $M$ is the number of classes, $y$ is the truth labels and $p(y_a)$ is the softmax  probability for $a^{th}$ class.  


The $\lambda$ in $F_{R}$ acts as a signal gate controller between the incoming signals from both CNN and the feature fusion component.  In other words, if $\lambda = 0$, then the signals pass the gate without constraints. However, if $\lambda$ is large, then the incoming signals from the fusion component have more freedom to pass the gate and feed into the CNN model with implicit interpretable features.

Therefore, we increase the feature reduction value to constrain the coming signals from the CNN model, which computes various kinds of features in an image such as edges, curves. This gives more freedom to approximated explanations from the explainable fusion component. Those features carry important explanations features of the teacher model. Then we merge the approximated SHAP explanation features of the teacher model with CNN features as follow:  

The operation of concatenation layer in the context of tensors, is the operation of joining tensors along one dimension using Linear layer. In particular, given $x \in \rm I\!R^{n x g}$ and $y \in \rm I\!R^{ m x g}$, where $x$ is CAE features, and $y$ is CNN features. The resulting tensor after concatenation $ca$ will then be:
\begin{align*}
    cat(x, y) = [x_1, ..., x_n, y_1, ..., y_m] \\
    \textrm{where}\; x_i \in \rm I\!R^{1 x g} \\
    \textrm{and}\; y_i \in \rm I\!R^{1 x g}
\end{align*}

Figure \ref{i_model} illustrates the details of XDistillation. It is crucial to note that in Xdistillation at the testing phase, we use the frozen weights of the CAE because the CAE trains separately and only once. The purpose behind using the pre-trained CAE and not train it with CNN model is that: (1) reducing the training time of using our model with different types of student models. (2) providing flexibility to our method to be used with various explainers without retraining the CAE. Training the explanations of only one explainer, SHAP values, for teacher model in Figure \ref{Fea_Fusion} and then use it in XDistillation  
improves the explanation output of the proposed model, which is similar to the actual explanations of the teacher model.

\section{Experimental Analysis} 
\label{exp_details}
In the following experiments, we analyze XDistillation in terms of predictive accuracy and explanation similarity (testing phase in Figure \ref{i_model}). In this section, we first describe our experimental setup and description of testing datasets. Next, we show the benefits of the explanation infusion step on the predictive performance of different models and compare XDistillation with other state-of-the-art distillation methods in term of test accuracy. Finally, we evaluate the consistency of the student model's explanations generated by different knowledge distillation methods.





\tabcolsep=0.10cm
\begin{table}[h]
  \caption{The hyper-parameters Details of the convolutional autoencoders}
  \label{cae_hp}
  \begin{tabular}{cccccl} 
    \toprule
    CAE\_Model & Type & Kernel size & Stride & Padding & \#parameters \\
    \midrule 
    \multicolumn{1}{r}{\hfil Small\_CAE}
    & Conv2d &  3&  &1 & \textbf{1077} \\
    & MaxPool2d &  2&  2& & \\
    & Conv2d &  3&  &1 & \\
    & MaxPool2d &  2&  2& & \\
    & ConvT &  2&  2& & \\
    & ConvT &  2&  2& & \\

 \midrule 
    \multicolumn{1}{r}{\hfil Large\_CAE}
    & Conv2d &  4&  2& 1& \textbf{198,796} \\ 
    & Conv2d &  4&  2& 1& \\
    & Conv2d &  4&  2& 1& \\
    & Conv2d &  4&  2& 1& \\
    & ConvT &  4&  2&1 & \\
    & ConvT &  4&  2&1 & \\
    & ConvT &  4&  2&1 & \\
    & ConvT &  4&  2&1 & \\

  \bottomrule
\end{tabular}
\end{table}

\renewcommand{\arraystretch}{1.1}
\begin{table*}[t]
  \centering
  \caption{Model Generalization}
  \label{Generalization}
  \begin{tabular}{ccccl} 
    \toprule
    Dataset & autoencoder &  Model & Accuracy $\%$ & \#parameters \\
    \midrule 
    \multicolumn{1}{r}{\hfil MNIST} 
    & \multicolumn{1}{r}{\hfil Small\_CAE} 
    &  Teacher(LeNet5) & 99&  61,701 \\
    &  &  Baseline student(Net) &97& 8,297  \\
    &  &  KD (Net) &\textbf{98}& 8,297 \\
    &  &  Xdistillation (Net) & \textbf{98}& 17,386  \\

 \midrule 
      \multicolumn{1}{r}{\hfil CIFAR-10} 
    & \multicolumn{1}{r}{\hfil Large\_CAE} 
    &  Teacher (VGG16)& 93.78 & 14,728, 266  \\
    &  &  Baseline student(VGG7) &89.2&  2,781,386  \\
    &  &  KD (VGG7) &90.2& 2,781,386\\
    &  &  Xdistillation (VGG7) &\textbf{90.9} & 3,521,276  \\
    
 \midrule 
      \multicolumn{1}{r}{\hfil CIFAR-10} 
    & \multicolumn{1}{r}{\hfil Large\_CAE} 
    &  Teacher(VGG16)& \textbf{93.78}& 14,728, 266  \\
    &  &  Baseline student (VGG11) &92&  9,231,114 \\
    &  &  KD (VGG11) &92.41& 9,231,114\\
    &  &  Xdistillation (VGG11) &\textbf{92.59}& 9,624,598  \\
  \bottomrule
\end{tabular}
\end{table*}

\subsection{Experiment Setting} \label{dataset}

\textbf{Dataset.} Our experiments are conducted on MNIST \cite{deng2012mnist} and CIFAR-10 \cite{krizhevsky2009learning} datasets. 
The MNIST dataset contains images of $28\times 28$ grayscale hand-written digits. In our experiments, the training-testing split is $60,000:10,000$. On the other hand, CIFAR-10 consists of $60,000$ $32\times 32$ colour images in 10 classes, with $6000$ images per class. There are $50,000$ training images and $10,000$ test images.

\textbf{Teacher Models.} In our experiments, the teacher models are implemented using LeNet5 \cite{lecun1998gradient} and VGG16 architecture\cite{ SimonyanZ14a} for the MNIST and CIFAR-10 dataset respectively. The LeNet is trained using MSE loss and Adam optimizer with learning rate $0.001$  and number of epochs $150$.
In CIFAR-10, our training included data augmentations with random horizontal flips and random crops of size $32$ with a possible padding of $4$ pixels. The training utilizes with a batch size of $128$ for $500$ epochs. The optimizer is stochastic gradient descent with a momentum of $0.9$ and a learning rate schedule of $0.1, 0.01$ and $0.001$ starting from the epochs $0$, $150$ and $250$ respectively. The models' accuracy and number of parameters are shown in Table \ref{Per_com}.

\textbf{Student Models.} The student models in our experiments are smaller LeNet5 for MNIST and VGG7\cite{ SimonyanZ14a} for CIFAR-10. Their testing accuracy and number of parameters can also be found in Table \ref{Per_com}. The student models are trained using some state-of-the-art distillation knowledge techniques, including knowledge distillation (KD) \cite{44873}, attention transfer (AT) \cite{zagoruyko2016paying}, neural selective transfer (NST) \cite{huang2017like}, and activation boundary (AB) \cite{heo2019knowledge}.

\textbf{XDistillation Model.} To train XDistillation, we first use SHAP \cite{lundberg2017unified}, LIME \cite{ribeiro2016should} and GradCAM \cite{Selvaraju_2017_ICCV} explainers to generate the teacher's explanations. Then, the explanations are transformed as described in section~\ref{shap_Rep} and the convolutional autoencoder as described in section \ref{cee_section} is trained based on these transformed explanations. The autoencoder is trained using the L1 loss and the Adam optimizer with a learning rate of $0.001$. The detail of the large CAE is shown in Table \ref{cae_hp}. The CNNs used in XDistillation have the same architecture and parameters as the student models. The CNN is trained using the loss Eq. (\ref{eq:total_loss}). We typically use $\alpha = 0.9$ and temperature $\tau = 1$. 

\begin{figure}[h]
\centering

    \includegraphics[width=0.8\columnwidth]{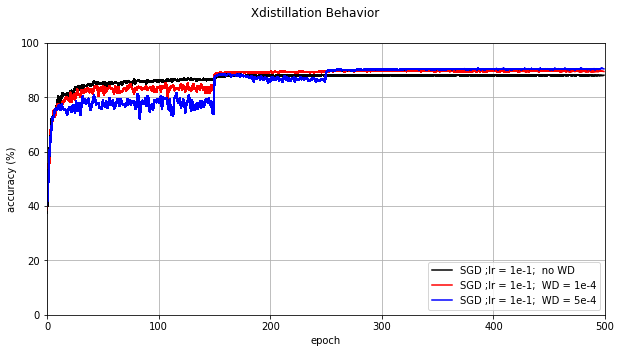} 
    \caption{ The testing accuracy of XDistillation during training with different regularized parameter $\lambda$ (equation \ref{eq:total_loss})}
     \label{abb}
\end{figure}

One key parameter that determines XDistillation's performance is the choice of the regularized parameter $\lambda$ controlling the model's signal from the CNN and the CAE as described in equation (\ref{eq:total_loss}). Figure \ref{abb} shows the accuracy of  XDistillation during training with different values of $\lambda$. In the following experiments, we use $\lambda = 5e-4$. 

\subsection{Results}
\label{Performance_Com}

\textbf{Predictive performance.} We first demonstrate the advantage of the explanation infusion step via CAE by comparing the test accuracy of XDistillation models with the baseline student models and the KD models. As shown in Table \ref{Generalization}, the introduction of the CAE generally increases the overall predictive accuracy of the student models.

Table \ref{Per_com} shows the predictive performance of the XDistillation model and compares it with other distillation knowledge methods. We can see that XDistillation outperform the other models in term of performance with an accuracy of $90.9\%$. The second close model to ours is the KD with $90.2\%$ accuracy, whereas the performance of the remaining models was below $90\%$. Generally, the accuracy improved in comparison to the baseline student model and other  state-of-the-art methods.



\renewcommand{\arraystretch}{1.1}
\begin{table}
  \centering
  \caption{Performance Comparison}
  \label{Per_com}
  \begin{tabular}{ccl} 
    \toprule
 Model & Accuracy $\%$ & \#parameters \\
 \midrule 
    Teacher & \textbf{93,78}& 14,728, 266  \\
        Baseline student&89.2&  2,781,386  \\
        Knowledge distillation (KD) &90.2& 2,781,386\\
         Attention transfer (AT)  &90& 2,781,386\\
         Neural selective transfer (NST) &89.47& 2,781,386\\
           Activation boundary (AB) &89.36& 2,781,386\\
    \midrule 
        Xdistillation  &\textbf{90.9} & 3,521,276  \\
    \bottomrule
    
\end{tabular}
\end{table}

\begin{figure}[h]
\centering
    \includegraphics[width=0.9\columnwidth]{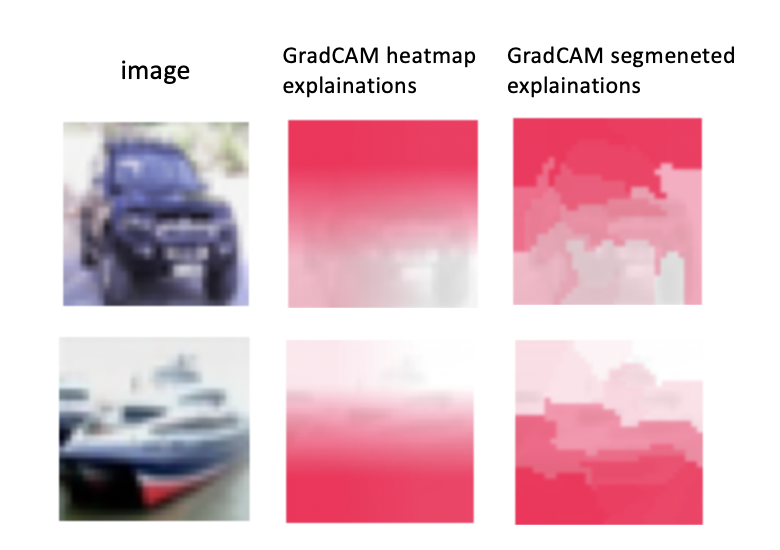}   
    \caption{GradCAM representations where the red color indicates the important features}
    \label{grad_heat}
\end{figure}

\renewcommand{\arraystretch}{1.2}
\begin{table}
  \caption{Scoring Distance}
  \label{distance_score}
  \centering
  \begin{tabular}{cccl} 
    \toprule
    \backslashbox{MSE}{Explainer} & SHAP & GradCAM & LIME   \\
    \midrule 
    KD & $0.027$&  $0.025$&\textbf{0.006}   \\
    AT & $0.040$ &  $0.239$& $0.014$ \\
    NST & $0.037$ &  $0.246$& $0.015$ \\
    AB & $0.033$ &  $0.199$& $0.014$ \\
    \midrule 
    \vspace{1pt}
    Xdistillation (our) & \textbf{0.026} &  \textbf{0.024}& \textbf{0.006} \\
    \bottomrule
\end{tabular}
\end{table}

\begin{table}
  \caption{Scoring explanations}
  \centering
  \label{overlap_comp}
  \begin{tabular}{cccl} 
    \toprule
    \backslashbox{Model}{Explainer} & SHAP & GradCAM & LIME   \\
    \midrule 
    KD & $68.3\%$ &  $56.1\%$ & $50.1\%$   \\
    AT & $67.9\%$ &  $52.1\%$& $44.3\%$  \\
    NST & $67.6\%$ &  $58.6\%$& $44.4\%$  \\
    AB & $67.3\%$ &  $55.6\%$& $44.7\%$  \\
    \midrule 
    \vspace{1pt}
    Xdistillation (our) & \textbf{70.5\%} &  \textbf{62\%}& \textbf{54.4\%}  \\

    \bottomrule
\end{tabular}
\end{table}




\textbf{Explanation consistency.} We now examine the ability of XDistillation model in providing consistent explainable features to the teacher model. Here, we use SHAP, GradCAM and, LIME explainers for our evaluation. Since there are some differences in the representation of the explainers' output. In particular, GradCAM proliferates as heatmap over an image, shown in Figure \ref{grad_heat}. On the other hand, LIME explainer utilizes quickshift segmentation approach \cite{vedaldi2008quick}. Thus, for a fair comparison, we apply the explanation's transformation step described in subsection \ref{shap_Rep} onto GradCAM and the SLIC segmentation procedure on LIME.

Given explanations of the same input of the same predictions from two models, we use the mean square error (MSE)~\cite{rezatofighi2019generalized} to measure the similarity between the two explanations. Table \ref{distance_score} reports the average MSE on $10,000$ pairs of explanations in CIFAR-10 experiments. In general, we can observe that the explanations of XDistillation are more similar to the teacher than other methods.


Since different explainers return different representations, beside MSE, we also measure the explanation consistency by measuring the overlapping area of explained super-pixels returned by each explainer. Specifically, for all correctly classified inputs by both teacher and student models, we collect the $k$ most important super-pixels of the corresponding explanations and compute the $sign(x)$ function as follows:
\begin{equation*}
    h(x) = \begin{cases}
    1 &\text{where} \; sign(M(x)) = sign(T(x))\\
    0 &\text{otherwise}\end{cases}
\end{equation*}
where $M(x)$ is the explanation value of overlapping SLIC super-pixels patch that classified correctly by the student model, $T(x)$ is the  explanation number of overlapping SLIC super-pixels patch that classified correctly by the teacher model, and $sign(.)$ is a function where it returns $1$  if the explanation segments of the teacher and the student have same sign. We then compare the counts of the correct  explanation patches $h(x)$ for both models, our XDistillation and other KD, based on the baseline model (VGG7), with normalization.

\begin{figure*}
\centering
     \includegraphics[width=1\textwidth]{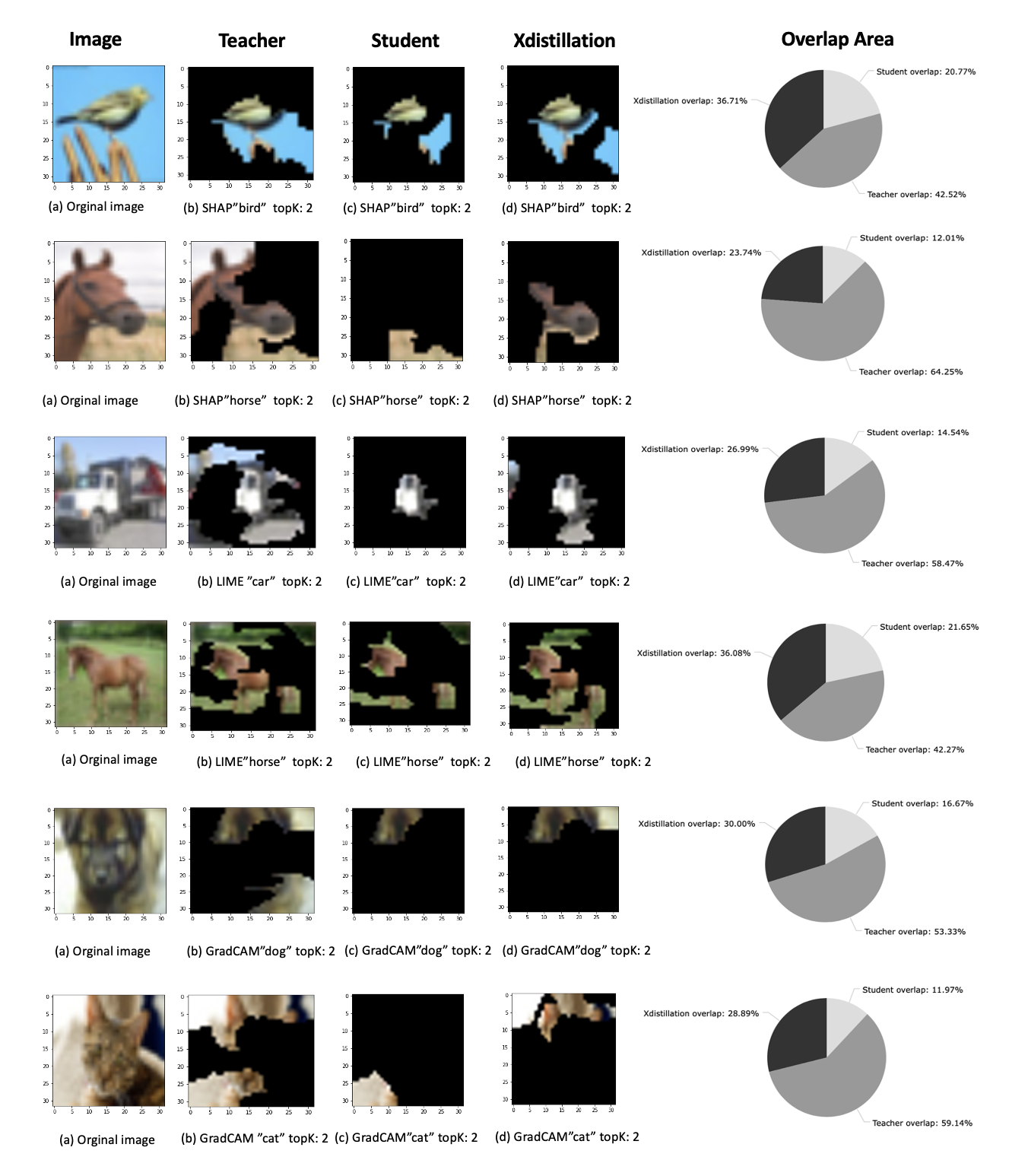} 
    \caption{We use SHAP, LIME and GradCAM to compare the overlap area of teacher, student and our proposed method to a original image for topK: 2 
 }
    \label{samples_explaintions}
\end{figure*}

Table \ref{overlap_comp} shows the overlapping-score of XDistillation and other methods. As can be seen, our proposed model has more similar explanations to the teacher model in comparison to knowledge distillation methods and baseline model. For instance, the overlap area of GradCAM for XDistillation method improved by about $6\%$ compare to the second close model to our method in term of the explanation performance which means our model is more similar to the teacher model. For reference, Figure \ref{samples_explaintions} provides examples of explanations of some predictions generated by SHAP, LIME and GradCAM along with the corresponding overlapping-scores.



\begin{figure}[h]
\centering
    \includegraphics[width=1\columnwidth]{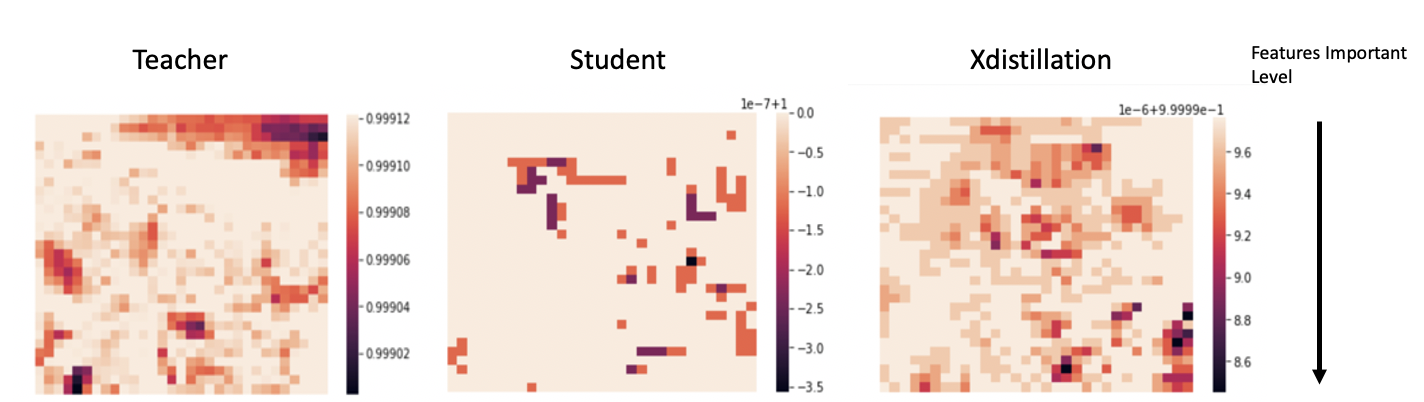}   
    \caption{Occlusion Analysis}
    \label{occ_ana}
\end{figure}

\textbf{Occlusion Analysis.}  Occlusion analysis is a simple method for understanding which parts of an image are most important for a model. Small perturbations of the data can be used to measure a network's sensitivity to occlusion in different regions of the data. The process for an occlusion experiment is as follows: we first mask part of an image before feeding it into three different trained models, including the (VGG16), our proposed model, and KD (VGG7) \cite{44873}. Then, we draw a heatmap of class scores for each masked image. Lastly, we slide the masked area to a different spot and repeat the step process again until we cover all the images. The reasoning behind all these is that if the class score for a partially occluded image is different than the true class, then the occluded area was likely very important.

Figure \ref{occ_ana} shows samples of the occlusion for the three models. In our human-subjective test, we apply the experiment for randomly 30 samples for each model whose predictions made {\em correctly}. thirty users were asked to choose which two of the three models are most likely similar to each other based on heatmap features. $18$ out of $30$ select the most likely models to each other are the teacher and XDistillation model where $5$ individuals choose XDistillation and KD model. We can observe that the the most important features captured by the two models, teacher and XDistillation, are most likely similar to each other.

\section{Conclusion} 
\label{sum_up}
Knowledge distillation (KD) tackles the issue of transferring knowledge from a vast and complicated neural network to a smaller one. The standard methodology reduces the KD divergence between a teacher and student model's outputs. However, current KD techniques ignore an important information of an explanatory network of the teachers. In this paper, we present XDistillation, explanatory model that mimics teacher explanations. We show experimentally that our proposed model outperforms the existing KD techniques. The Xdistillation method offers the ability to eliminate incoherence of the explanations between the teacher and the student patterns apart from current KD strategies.

\section*{Acknowledgment}

This work was supported in part by the National Science Foundation Program on Fairness in AI in collaboration with Amazon under award No. 1939725.


\bibliographystyle{IEEEtran}
\bibliography{Raedbibfile}

\begin{thebibliography}{10}
\providecommand{\url}[1]{#1}
\csname url@samestyle\endcsname
\providecommand{\newblock}{\relax}
\providecommand{\bibinfo}[2]{#2}
\providecommand{\BIBentrySTDinterwordspacing}{\spaceskip=0pt\relax}
\providecommand{\BIBentryALTinterwordstretchfactor}{4}
\providecommand{\BIBentryALTinterwordspacing}{\spaceskip=\fontdimen2\font plus
\BIBentryALTinterwordstretchfactor\fontdimen3\font minus
  \fontdimen4\font\relax}
\providecommand{\BIBforeignlanguage}[2]{{%
\expandafter\ifx\csname l@#1\endcsname\relax
\typeout{** WARNING: IEEEtran.bst: No hyphenation pattern has been}%
\typeout{** loaded for the language `#1'. Using the pattern for}%
\typeout{** the default language instead.}%
\else
\language=\csname l@#1\endcsname
\fi
#2}}
\providecommand{\BIBdecl}{\relax}
\BIBdecl

\bibitem{44873}
G.~Hinton, O.~Vinyals, and J.~Dean, ``Distilling the knowledge in a neural
  network,'' in \emph{NIPS Deep Learning and Representation Learning Workshop},
  2015.

\bibitem{han2015learning}
S.~Han, J.~Pool, J.~Tran, and W.~J. Dally, ``Learning both weights and
  connections for efficient neural network,'' in \emph{NIPS}, 2015.

\bibitem{cho2019efficacy}
J.~H. Cho and B.~Hariharan, ``On the efficacy of knowledge distillation,'' in
  \emph{Proceedings of the IEEE/CVF International Conference on Computer
  Vision}, 2019, pp. 4794--4802.

\bibitem{alharbi2021evaluating}
R.~Alharbi, M.~N. Vu, and M.~T. Thai, ``Evaluating fake news detection models
  from explainable machine learning perspectives,'' in \emph{ICC 2021-IEEE
  International Conference on Communications}.\hskip 1em plus 0.5em minus
  0.4em\relax IEEE, 2021, pp. 1--6.

\bibitem{arrieta2020explainable}
A.~B. Arrieta, N.~D{\'\i}az-Rodr{\'\i}guez, J.~Del~Ser, A.~Bennetot, S.~Tabik,
  A.~Barbado, S.~Garc{\'\i}a, S.~Gil-L{\'o}pez, D.~Molina, R.~Benjamins
  \emph{et~al.}, ``Explainable artificial intelligence (xai): Concepts,
  taxonomies, opportunities and challenges toward responsible ai,''
  \emph{Information Fusion}, vol.~58, pp. 82--115, 2020.

\bibitem{Geirhos2020}
R.~Geirhos, J.-H. Jacobsen, C.~Michaelis, R.~Zemel, W.~Brendel, M.~Bethge, and
  F.~A. Wichmann, ``Shortcut learning in deep neural networks,'' \emph{Nature
  Machine Intelligence}, vol.~2, no.~11, Nov 2020.

\bibitem{lundberg2017unified}
S.~M. Lundberg and S.-I. Lee, ``A unified approach to interpreting model
  predictions,'' in \emph{NIPS}, 2017.

\bibitem{zagoruyko2016paying}
N.~Komodakis and S.~Zagoruyko, ``Paying more attention to attention: improving
  the performance of convolutional neural networks via attention transfer,'' in
  \emph{ICLR}, 2017.

\bibitem{heo2019knowledge}
B.~Heo, M.~Lee, S.~Yun, and J.~Y. Choi, ``Knowledge transfer via distillation
  of activation boundaries formed by hidden neurons,'' in \emph{Proceedings of
  the AAAI Conference on Artificial Intelligence}, vol.~33, no.~01, 2019, pp.
  3779--3787.

\bibitem{tung2019similarity}
F.~Tung and G.~Mori, ``Similarity-preserving knowledge distillation,'' in
  \emph{Proceedings of the IEEE/CVF International Conference on Computer
  Vision}, 2019, pp. 1365--1374.

\bibitem{vu2020pgm}
M.~Vu and M.~T. Thai, ``Pgm-explainer: Probabilistic graphical model
  explanations for graph neural networks,'' \emph{Advances in Neural
  Information Processing Systems}, vol.~33, pp. 12\,225--12\,235, 2020.

\bibitem{lipton2018mythos}
Z.~C. Lipton, ``The mythos of model interpretability: In machine learning, the
  concept of interpretability is both important and slippery.'' \emph{Queue},
  vol.~16, no.~3, pp. 31--57, 2018.

\bibitem{zhou2016learning}
B.~Zhou, A.~Khosla, A.~Lapedriza, A.~Oliva, and A.~Torralba, ``Learning deep
  features for discriminative localization,'' in \emph{Proceedings of the IEEE
  conference on computer vision and pattern recognition}, 2016, pp. 2921--2929.

\bibitem{Selvaraju_2017_ICCV}
R.~R. Selvaraju, M.~Cogswell, A.~Das, R.~Vedantam, D.~Parikh, and D.~Batra,
  ``Grad-cam: Visual explanations from deep networks via gradient-based
  localization,'' in \emph{Proceedings of the IEEE International Conference on
  Computer Vision (ICCV)}, Oct 2017.

\bibitem{fong2019understanding}
R.~Fong, M.~Patrick, and A.~Vedaldi, ``Understanding deep networks via extremal
  perturbations and smooth masks,'' in \emph{Proceedings of the IEEE/CVF
  International Conference on Computer Vision}, 2019, pp. 2950--2958.

\bibitem{ribeiro2016should}
M.~T. Ribeiro, S.~Singh, and C.~Guestrin, ``" why should i trust you?"
  explaining the predictions of any classifier,'' in \emph{Proceedings of the
  22nd ACM SIGKDD international conference on knowledge discovery and data
  mining}, 2016, pp. 1135--1144.

\bibitem{wang2016auto}
Y.~Wang, H.~Yao, and S.~Zhao, ``Auto-encoder based dimensionality reduction,''
  \emph{Neurocomputing}, vol. 184, pp. 232--242, 2016.

\bibitem{lore2017llnet}
K.~G. Lore, A.~Akintayo, and S.~Sarkar, ``Llnet: A deep autoencoder approach to
  natural low-light image enhancement,'' \emph{Pattern Recognition}, vol.~61,
  pp. 650--662, 2017.

\bibitem{chen2017outlier}
J.~Chen, S.~Sathe, C.~Aggarwal, and D.~Turaga, ``Outlier detection with
  autoencoder ensembles,'' in \emph{Proceedings of the 2017 SIAM international
  conference on data mining}.\hskip 1em plus 0.5em minus 0.4em\relax SIAM,
  2017, pp. 90--98.

\bibitem{masci2011stacked}
J.~Masci, U.~Meier, D.~Cire{\c{s}}an, and J.~Schmidhuber, ``Stacked
  convolutional auto-encoders for hierarchical feature extraction,'' in
  \emph{International conference on artificial neural networks}.\hskip 1em plus
  0.5em minus 0.4em\relax Springer, 2011, pp. 52--59.

\bibitem{ashfahani2020devdan}
A.~Ashfahani, M.~Pratama, E.~Lughofer, and Y.-S. Ong, ``Devdan: Deep evolving
  denoising autoencoder,'' \emph{Neurocomputing}, vol. 390, pp. 297--314, 2020.

\bibitem{SimonyanZ14a}
K.~Simonyan and A.~Zisserman, ``Very deep convolutional networks for
  large-scale image recognition,'' in \emph{3rd International Conference on
  Learning Representations, {ICLR} 2015, San Diego, CA, USA, May 7-9, 2015,
  Conference Track Proceedings}, Y.~Bengio and Y.~LeCun, Eds., 2015.

\bibitem{krizhevsky2009learning}
A.~Krizhevsky, G.~Hinton \emph{et~al.}, ``Learning multiple layers of features
  from tiny images,'' 2009.

\bibitem{noh1976slic}
W.~F. Noh and P.~Woodward, ``Slic (simple line interface calculation),'' in
  \emph{Proceedings of the fifth international conference on numerical methods
  in fluid dynamics June 28--July 2, 1976 Twente University, Enschede}.\hskip
  1em plus 0.5em minus 0.4em\relax Springer, 1976, pp. 330--340.

\bibitem{ranzato2007unsupervised}
M.~Ranzato, F.~J. Huang, Y.-L. Boureau, and Y.~LeCun, ``Unsupervised learning
  of invariant feature hierarchies with applications to object recognition,''
  in \emph{2007 IEEE conference on computer vision and pattern
  recognition}.\hskip 1em plus 0.5em minus 0.4em\relax IEEE, 2007, pp. 1--8.

\bibitem{NIPS2017_7062}
\BIBentryALTinterwordspacing
S.~M. Lundberg and S.-I. Lee, ``A unified approach to interpreting model
  predictions,'' in \emph{Advances in Neural Information Processing Systems
  30}, I.~Guyon, U.~V. Luxburg, S.~Bengio, H.~Wallach, R.~Fergus,
  S.~Vishwanathan, and R.~Garnett, Eds.\hskip 1em plus 0.5em minus 0.4em\relax
  Curran Associates, Inc., 2017, pp. 4765--4774. [Online]. Available:
  \url{http://papers.nips.cc/paper/7062-a-unified-approach-to-interpreting-model-predictions.pdf}
\BIBentrySTDinterwordspacing

\bibitem{deng2012mnist}
L.~Deng, ``The mnist database of handwritten digit images for machine learning
  research,'' \emph{IEEE Signal Processing Magazine}, vol.~29, no.~6, pp.
  141--142, 2012.

\bibitem{lecun1998gradient}
Y.~LeCun, L.~Bottou, Y.~Bengio, and P.~Haffner, ``Gradient-based learning
  applied to document recognition,'' \emph{Proceedings of the IEEE}, vol.~86,
  no.~11, pp. 2278--2324, 1998.

\bibitem{huang2017like}
Z.~Huang and N.~Wang, ``Like what you like: Knowledge distill via neuron
  selectivity transfer,'' \emph{arXiv preprint arXiv:1707.01219}, 2017.

\bibitem{vedaldi2008quick}
A.~Vedaldi and S.~Soatto, ``Quick shift and kernel methods for mode seeking,''
  in \emph{European conference on computer vision}.\hskip 1em plus 0.5em minus
  0.4em\relax Springer, 2008, pp. 705--718.

\bibitem{rezatofighi2019generalized}
H.~Rezatofighi, N.~Tsoi, J.~Gwak, A.~Sadeghian, I.~Reid, and S.~Savarese,
  ``Generalized intersection over union: A metric and a loss for bounding box
  regression,'' in \emph{Proceedings of the IEEE/CVF Conference on Computer
  Vision and Pattern Recognition}, 2019, pp. 658--666.

\end{thebibliography}
\end{document}